# 基于语义–表观特征融合的无监督前景分割

李熹 [a]，马惠敏 [b,*]，马洪兵 [a,c]，王弈冬 [a]

a. 清华大学，北京市，100084； b. 北京科技大学，北京市，100083； c. 新疆大学，新疆省，830046

**摘 要**：前景分割是图像理解领域中的重要任务，在无监督条件下，由于不同图像、不同实例往往具有多变的表达形式，这使得基于固定规则、单一类型特征的方法很难保证稳定的分割性能。针对这一问题，本研究提出了一种基于语义-表观特征融合的无监督前景分割方法（SAFF）。在研究中发现，语义特征能够对前景物体关键区域产生精准的响应，而以显著性、边缘为代表的表观特征则提供了更丰富的细节表达信息。为了融合了二者的优势，研究建立了融合语义、表观信息的一元区域特征和二元上下文特征编码的方法，实现了对两种特征表达的全面描述。接着，一种图内自适应参数学习的方法被设计，用于计算最适合的特征权重，并生成前景置信分数图。进一步的，分割网络被使用来学习不同实例间前景的共性特征。通过融合语义和表观特征，并级联图内自适应特征权重学习、图间共性特征学习的模块，研究在 PASCAL VOC 2012 数据集上取得了显著超过基线的前景分割性能。

**关键词**：前景分割；无监督学习；语义-表观特征融合；自适应加权

## Unsupervised segmentation via semantic-apparent feature fusion

Xi Li[a], Huimin Ma[b,*], Hongbing Ma[a,c], Yidong Wang[a]

a. Tsinghua University, Beijing, 100084

b. University of Science and Technology Beijing, Beijing, 100083

c. Xinjiang University, Xinjiang, 830046

**Abstract:** Foreground segmentation is an essential task in the field of image understanding. Under unsupervised conditions, different images and instances always have variable expressions, which make it difficult to achieve stable segmentation performance based on fixed rules or single type of feature. In order to solve this problem, the research proposes an unsupervised foreground segmentation method based on semantic-apparent feature fusion (SAFF). Here, we found that key regions of foreground object can be accurately responded via semantic features, while apparent features (represented by saliency and edge) provide richer detailed expression. To combine the advantages of the two type of features, an encoding method for unary region features and binary context features is established, which realizes a comprehensive description of the two types of expressions. Then, a method for adaptive parameter learning is put forward to calculate the most suitable feature weights and generate foreground confidence score map. Furthermore, segmentation network is used to learn foreground common features from different instances. By fusing semantic and apparent features, as well as cascading the modules of intra-image adaptive feature weight learning and inter-image common feature learning, the research achieves performance that significantly exceeds baselines on the PASCAL VOC 2012 dataset.

**Key words:** Foreground segmentation; unsupervised learning; semantic-apparent feature; adaptive weighting

## 1　研究背景

前景分割是图像目标检测[1, 2]、语义分割[3, 4]、显著性检测[5, 6]等领域中重要的任务之一。给定一张图像，前景分割的任务目标是提供一种实现像素级前景、背景分类的模型。在现有的研究中，基于全监督学习的方法[3, 7]已

经取得了良好的性能。然而，在很多实际情况中，经常面临着精确的像素级标注难以大量、高效获取的问题。因此，在弱监督、无监督条件下实现前景分割和语义分割，成为一项具有重要意义的工作。

本研究面向无监督前景分割任务，提出了一种语义-表观特征融合的分割方法（SAFF）。研究观察到，图像中的高维语义特征往往能够实现对前景关键区域的捕捉，但缺乏细节信息；而以显著性、边缘等为代表的表观特征，则更好的提供了图像细节表达和对局部区域间相似关系的描述，但缺乏对前景物体的语义表达。同时，对于不同的图像、实例而言，其语义特征和表观特征具有不同的表达能力和精度。因此，本研究建立了一种自适应参数学习的方式，提出了一种对每张图像语义和表观特征的最优加权策略，实现了对前景区域的精确提取。接着，多实例学习的框架被使用来实现不同图像前景共性特征的计算，并进一步实现了对图像前景分割结果的重推理和性能优化。

在已有研究中，基于弱监督和全监督的语义分割方法为本研究提供了参考。一方面，现有面向前景及语义分割任务的研究，提供了多种全监督条件的学习框架。以全卷积网络（FCN）[7]为基础，DeepLab [3]及其变体通过引入了空洞卷积[8, 9]、条件随机场（CRFs）[10]、深度空间金字塔池化（ASPP）等单元，得到了更高的性能。而针对无监督任务而言，虽然基于全监督的深度学习框架不能直接实现对像素级分割的学习与推理，但其仍然可以被利用来学习不同图像和实例间的共性表达。在本研究中，基于大量图像样本和对应的初始分割结果，全监督框架被使用来实现对前景共性特征的学习。进一步，利用训练好的分割网络，每一张图像可以被重新推理得到性能更高的分割结果。

另一方面，当前弱监督条件下的前景或语义分割任务得到了广泛的关注。其中，只利用图像标签的弱监督条件最具有难度。在这一设置下，采用两阶段式的方法取得了更高的性能，即首先生成伪标签信息，再利用全监督框架实现语义分割过程[4, 11, 12, 13]。在伪标签生成中，类别激活映射（CAM）模型[14]被广泛用来计算图像对分类贡献具有高响应的语义区域。基于初始定位种子，AffinityNet [11]提出了一种从伪标签中学习像素间相似关系的方法，实现对初始标注精度的提升；DSRG [12]，MCOF [4]，WRSS [13]则分别考虑在全监督训练过程中利用不同的迭代结构提升分割的性能。受到这些工作启发，本研究同样采用两阶段的方式实现前景分割任务。对于一张图像而言，由于其只包含前景和背景两个类别，因此无监督前景分割任务可以被看做弱监督语义分割的一种变体。本研究首先建立了基于图内上下文信息的前景学习模型，进一步将生成的特征图作为伪标签信息，利用分割网络实现图间多实例学习过程，最终得到高质量的分割结果。

本文的贡献主要包括以下三个方面：

其一，提出了一种语义和表观特征联合编码的无监督图像前景分割方法，并建立了一种图像内上下文相似度计算模型，得到了对图像前景信息的判别性表达。

其二，提出了一种自适应权重的图像内区域前景置信度学习和推理方法，自适应地实现对不同图像、实例的最优特征加权，实现了高质量的分割结果。分割网络被级联使用来学习多实例间的共性特征，并进一步提升了性能。

其三，针对无监督前景分割任务，本研究在 PASCAL VOC 2012 数据集上得到的分割结果取得了明显高于基线的性能。

## 2　算法框架

给定一张图像，本研究首先将其编码为语义特征和表观特征。具体的，研究用在 ImageNet [15]上预训练的特征提取器作为高维语义编码模块，用显著性、边缘特征编码图像的表观信息。同时，图像的超像素分割结果将被计算。对于每个超像素，研究中都以 4 维特征来描述其前景置信度。具体的，对于研究中首先利用语义特征各维度的最大响应值和显著性值来分别描述该超像素区域的一元区域语义、表观特征。接着，研究定义了两两超像素之间的语义特征相似度和边缘连通度，结合图像一元超像素区域特征，计算生成了二元上下文语义和表观特征。这里，语义特征相似度、边缘连通度的计算参考了工作[4, 5]。

进一步，本研究提出了一种自适应权重计算的方法。对于每个超像素，以这 4 维特征作

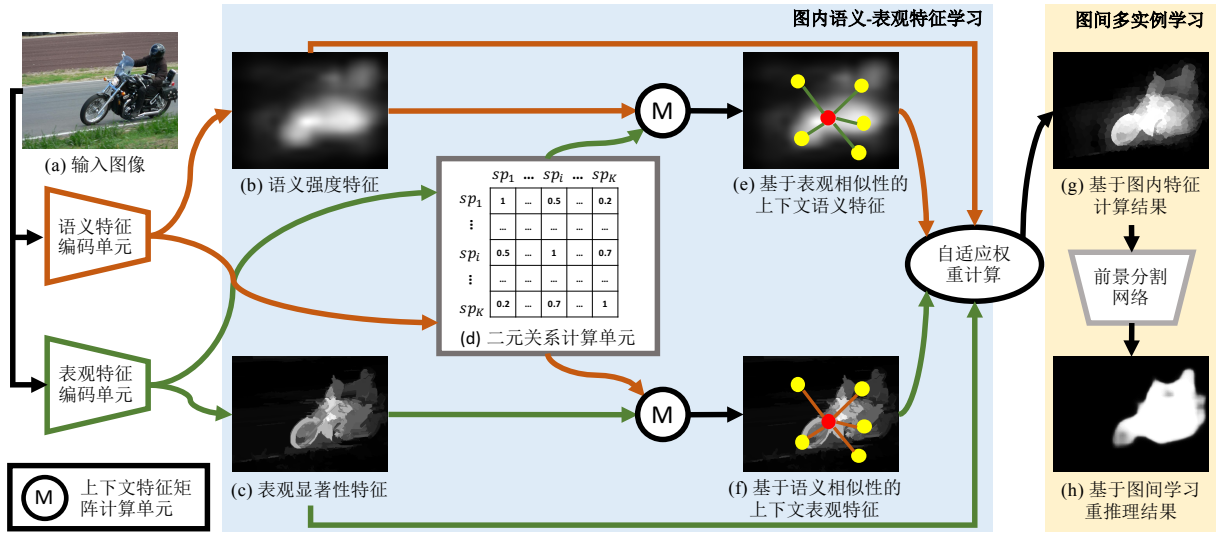

图 1 基于语义-表观特征融合的前景分割算法（SAFF）框架

为输入，通过自适应学习得到的权重，实现对前景置信度的推理。进一步的，对于 N 张图像样本，N 个对应的前景分割结果将被得到。此时，将这些前景分割结果作为伪标签，本研究采用了全监督分割网络的框架，实现了对图像多实例间前景共性特征的学习。基于训练完成的模型，不仅这 N 张用于训练的图像样本可以被重推理优化，得到更精确的前景分割结果，其他的图像也可以直接被推理实现前景分割。

图 1 展示了本研究提出的基于语义-表观特征融合的前景分割算法完整框架。

## 3 语义-表观特征编码

### 3.1 特征图生成及预处理

为了编码图像的表观特征和语义特征，本研究首先计算了图像的语义特征响应图、显著性目标图和边缘图。一方面，对于语义特征的生成，由于无监督条件下没有针对目标数据集的额外标注信息，研究采用基于 ImageNet [15] 预训练的类别激活映射（CAM）[14]模型生成特征图。这里，GoogleNet [16]被作为基础特征提取框架。对于一张图像，1000 维的响应特征将编码得到。需要说明，研究只将该模型作为语义特征提取器，并没有使用 ImageNet 的类别信息，，也没有使用训练集中的其它标注信息。另一方面，显著性响应图和边缘特征图被使用作为表观特征。特别的，本研究采用了 DRFI [17] 方法进行显著性特征的计算，而 EdgeBoxes [18] 方法则被用于提取边缘信息。需要说明的是，任何一种高维语义热图、显著性、边缘的特征编码算法都可以用于语义和表观特征计算，即研究提出的方法可以广泛的将不同模型得到的特征采用作为初始信息，并进一步实现对图像中前景区域的更精准分割。

同时，研究中将图像进行了超像素分割，并以超像素代表图像中最小的计算单元，以同时提高局部特征的表达能力和保证计算的高效性。利用特征编码结果，对于每个超像素$sp_i$，研究定义了语义、表观特征为其包含的所有像素对应特征在欧式空间的均值。特别的，$vs_i$表示该超像素的 1000 维归一化语义表达，且各维度的值分布在[0,1]之间；$S_a(i)$则表示其归一化的显著性特征。参考 AO-AIM [5]方法的定义，$E(i,j)$被定义为超像素$sp_i$和$sp_j$之间的边缘重量。其值非负，且越大的值代表了超像素对之间拥有越显著、越丰富的边缘表达。

### 3.2 一元区域特征编码

基于语义、表观特征，研究首先针对每个超像素区域进行直接地一元区域特征编码。特别的，在生成每种一元区域特征和二元上下文特征时，研究中都将其值编码在[0,1]的区间。具体的，对于一个超像素$sp_i$，其一元区域语义特征定义为$vs_i$各个维度中的最大值$S_s(i)$，表观特征则直接以显著性特征$S_a(i)$为代表。对于一张包含 K 个超像素的样本而言，2 个 K 维的特征向量$[...,S_s(i),...]^T$和$[...,S_a(i),...]^T$可以被利用来分别描述整张图像的一元语义和表观特征。

### 3.3 二元上下文特征编码

利用语义特征表达，对于一张图像中的两两超像素对$sp_i$和$sp_j$，研究用式(1)定义了该超像素对语义相似度的表达：

$$M_s(i,j) = \sum_{d=1}^{1000} min\,(vs_{i,d}, vs_{j,d}) \qquad (1)$$

另一方面，参考了[5, 10]等工作，研究中用式(2)来定义超像素对之间的表观相似度：

$$M_a(i,j) = exp\,(-w_e \cdot E(i,j)) \qquad (2)$$

这里，$w_e$是一个正数，在实验中，研究参考了工作[5, 13]，将其值设置为3.5。容易得到，$M_s(i,j)$和$M_a(i,j)$都是分布在[0,1]上的测度，且越大的值对应了两者之间具有越高的特征相似度或越小的边缘图上距离。

利用基于语义相似度和表观相似度的计算方法，图像中两两超像素对之间的相似度量可以表征为矩阵形式$M_s$，$M_a$。进一步的，我们将其对角元置零，并对矩阵逐行进行加和归一化处理，得到修正后的矩阵$M_s{'}$，$M_a{'}$。对于每一个超像素$sp_i$，修正后的矩阵$M_*{'}$可以看做其他所有超像素与该超像素的相似度量为归一化权重，故研究中以加权平均的方式，定义了图中每个超像素的语义和表观上下文特征$S_{s_{ctx}}(i)$、$S_{a_{ctx}}(i)$。式(3)和(4)给出了矩阵形式的全部超像素的上下文特征计算表达方法：

$$[\dots, S_{a_{ctx}}(i), \dots]^T = M_s{'} \cdot [\dots, S_a(i), \dots]^T \quad (3)$$

$$[\dots, S_{s_{ctx}}(i), \dots]^T = M_a{'} \cdot [\dots, S_s(i), \dots]^T \quad (4)$$

需要指出，在计算二元特征时，表观上下文特征是基于语义相似度矩阵和一元区域表观特征计算得到的；而语义上下文特征的生成则参考了表观相似度量和一元语义特征。这种交叉特征推理的方式，一方面，更好的融合了两种特征并为前景区域推理提供了更丰富的表达；另一方面，也避免了同一种特征同时被应用于一元表达和二元推理过程造成的误差累积。

## 4 自适应权重学习与推理

### 4.1 自适应权重学习

基于得到的一元、二元语义-表观编码特征，本研究使用加权求和的方式，生成前景置信度图。对于每一个超像素$sp_i$，其置信分数$Sco_{saff}(i)$通过式(5)定义：

$$Sco_{saff}(i) = Sco_v(i)^T \cdot w + bias \qquad (5)$$

其中，$w$为编码特征权重向量，$bias$为偏置系数，$Sco_v(i)$代表了该超像素编码得到的4元编码特征，如式(6)：

$$Sco_v(*) = [S_s,\ S_{s_{ctx}},\ S_a,\ S_{a_{ctx}}]^T \qquad (6)$$

考虑到对于不同的图像而言，一元、二元语义特征和表观特征对前景推理的影响不同，因此，研究中并不把$w$和$bias$设置成固定参数，而是采用自适应的方式实现参数的选取。首先，研究基于先验知识，假设当超像素的一元区域的表观、语义特征均具有高前景置信分数时应为前景，而均具有低置信分数时应判别为背景，并由此生成伪标注信息。具体的，给定一张图像，对于每一个超像素$sp_i$，研究中计算它的一元表观特征和语义特征的几何平均值，并分别选取值小于$th_{bg}$和大于$th_{fg}$的超像素，将其置信分数设置为0和1作为伪标签。(在实验中，$th_{bg}$和$th_{fg}$被分别设置为0.2和0.6)。此时，将每一组结果带入式(5)，可以得到以$w$和$bias$为自变量的超定方程。此时，公式(5)可以被作为定义了左右两端变量的目标函数，看做线性回归问题。这里，最小二乘方法被使用，来得到对线性条件下最优参数的计算。在得到了权重$w$和偏置$bias$后，式(5)可以被再次使用，对图像中每一个超像素的前景置信度进行推理。特别的，这里限制分数的范围是[0,1]的区间，对于在区间外的结果，研究中将其映射到相近的边界上。

需要额外说明的是，在生成伪标签时，研究观察到一张图像中选取出的背景样本往往远多于前景样本，而这会使得参数估计时对前背景置信分数计算产生一定偏差。因此，研究采用了一种样本均衡策略，即对较少的一类样本进行重采样，再将前背景平衡后的样本用于自适应参数估计的计算中。最终，将超像素映射会图像对应位置，可以得到一张对应的值分布在[0,1]区间的前景置信度图。

### 4.2 基于分割网络的重推理优化

对于每一张图像，研究提出的模型可以获取到融合了图像内部语义和表观上下文先验的前景区域推理结果。进一步地，基于$N$张包含了不同场景和实例的图像，研究利用全监督前景分割网络学习的框架，实现对不同图像、多种实例中前景的共性特征表达学习。首先，研

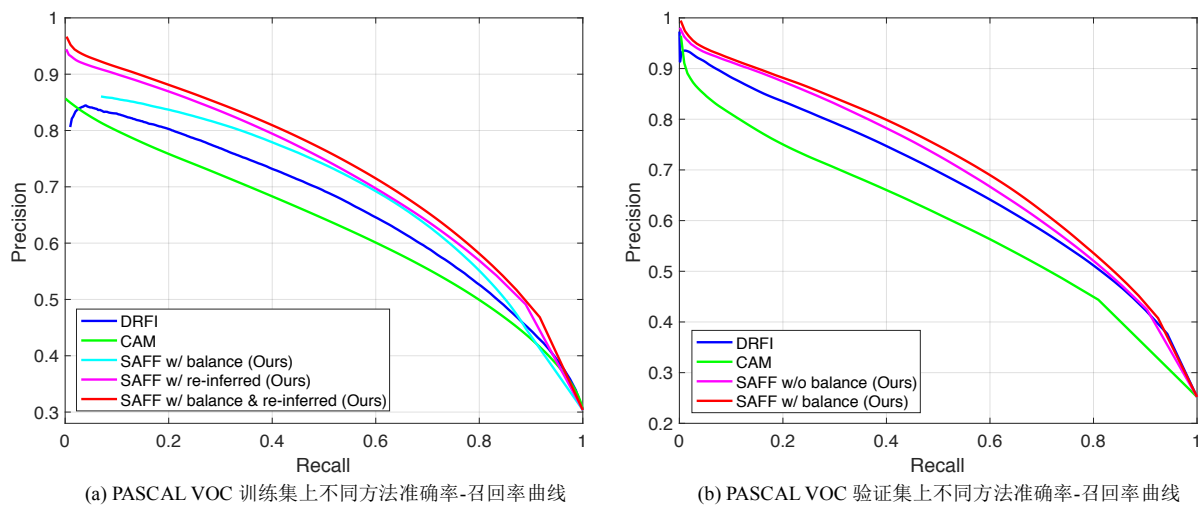

(a) PASCAL VOC 训练集上不同方法准确率-召回率曲线　　(b) PASCAL VOC 验证集上不同方法准确率-召回率曲线

图 2 不同方法在数据集上的准确率-召回率曲线性能比较

究将每张图像对应的前景分割置信度图进行二值化操作,以 0.5 为阈值划分前景和背景,生成 N 张伪标签。利用伪标签,研究可以基于全监督学习的方式实现对分割网络的训练。在本研究中,以 VGG-16 [19]为基础网络的 DeepLab-V2 [3]模型被采用,事实上,该模块可以被任意语义分割或前景分割模型代替。利用训练完成的模型,每一张图像可以被重新推理,并生成新的前景分割置信度图。同时,训练得到的网络模型也可以被直接对其他图像进行前景分割的推理。通过融合基于图内语义-表观信息的上下文学习,和基于图间前景共性特征的多实例学习,更好的前景分割结果可以被获得。

## 5　实验

### 5.1　数据集及评估方法设置

本文将提出的算法基于 PASCAL VOC 2012 图像分割数据集[20]进行性能评估。特别的,本研究中使用了增广数据[21],使得训练集中共包含 10,582 张图像样本。这一数据集包括了 20 类前景物体和 1 类背景,在评估时,本研究将所有属于物体类别的像素设置为前景。

本研究基于准确率($Precision$)、召回率($Recall$)和 F-测度指标($F_{measure}$)对前景分割的性能进行了评估。具体地,对于每一张图像和对应的前景置信图,研究将归一化置信度热图映射到从 0 到 255 的区间,并分别计算以 0 到 255,共 256 阈值划分下提取出的前景区域像素,分别计算准确率和召回率。进一步的,

F-测度指标可以被得到。F-测度的定义如式(7):

$$F_{measure} = \frac{(1+\beta^2)\cdot Precision \times Recall}{\beta^2 \cdot Precision + Recall} \qquad (7)$$

这里,参考典型的显著性区域检测、前景分割的方法[5, 22],本研究中将$\beta^2$设置为 0.3(在评估时,所有组中结果的最大值被记录)。

### 5.2　实验性能及分析

针对 PASCAL VOC 2012 的训练集,研究通过准确率-召回率曲线和 F-测度指标的对比来展示方法的性能。同时,为了比较本研究方法的性能并证明各个步骤的有效性,基线方法和本算法中各个模块分步优化的结果都被进行了评估。具体的,基于 DRFI [17]显著性物体检测、基于 CAM [14]的语义热图生成,及基于这些方法生成的伪标签用于分割网络训练并重推理优化的结果以基线被陈列;同时,本研究提出的 SAFF 方法在样本均衡处理、重推理优化模块独立、联合使用的性能也被评估。需要指出,本研究针对的是无监督条件的前景分割问题,因此并没有采用有专门任务标签训练条件下生成的语义和表观特征的方法作为基线模块。事实上,任何一种语义特征和表观特征编码方法,都可以用于本研究提出的学习结构,实现对前景分割任务性能的优化。

表 1 和表 2 分别展示了不同方法在 PASCAL VOC 2012 训练集和验证集上的 F-测度指标比较。需要指出,在验证集上的结果是由以不同方法为伪标签训练的分割网络直接推理得到的,并未对验证集图像进行其他操作。因此,基于本研究方法训练得到的模型进行前景分割的时

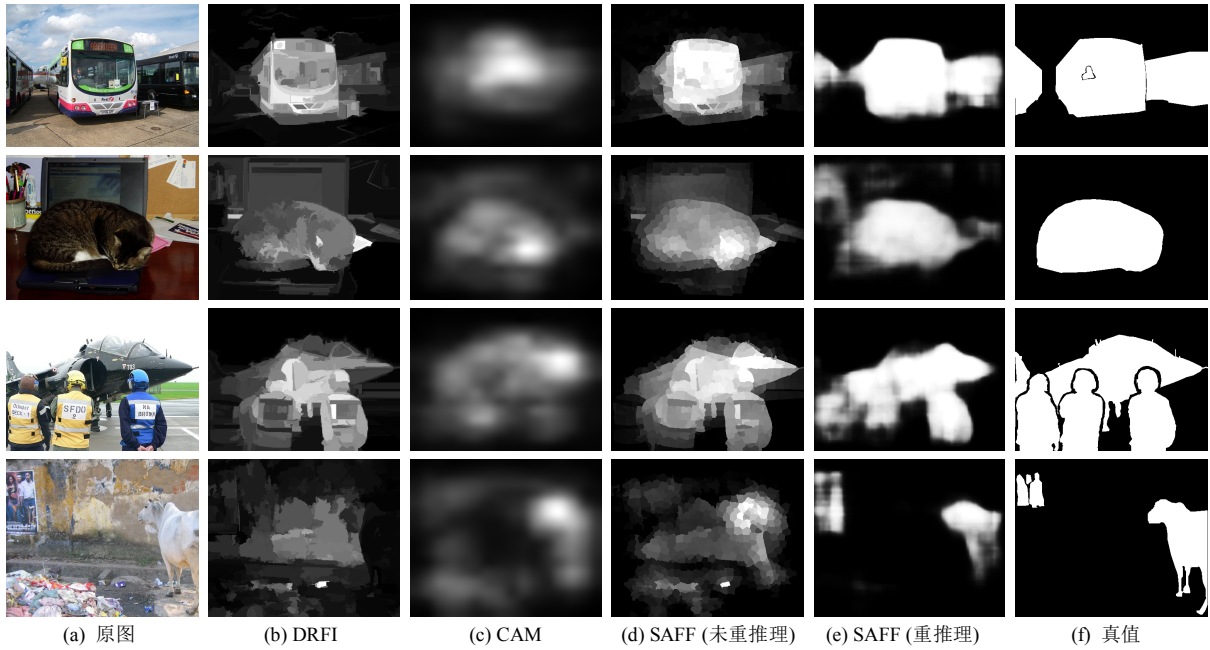

(a) 原图　　(b) DRFI　　(c) CAM　　(d) SAFF (未重推理)　　(e) SAFF (重推理)　　(f) 真值

图 3 不同方法得到的可视化结果对比。

间开销仅包括所使用的端到端分割网络推理时间，可以高效地应用于实际分割任务。能够看到，本研究提出的方法通过自适应的调整语义特征和表观特征对前景置信分数计算的贡献，实现了更高性能的前景分割。另外，在验证集上的结果也证明了基于 SAFF 得到的前景分割确率-召回率曲线的评估结果同样证明了本研究提出的方法各个模块性能的有效性。

表 1 PASCAL VOC 2012 训练集前景分割 F-测度比较

| 方法 | 样本均衡 | 网络重推理 | F-测度 |
|---|---|---|---|
| DRFI [16] | | | 0.6383 |
| DRFI [16] | | √ | 0.6513 |
| CAM [14] | | | 0.6044 |
| CAM [14] | | √ | 0.6116 |
| SAFF (Ours) | | | 0.6635 |
| SAFF (Ours) | √ | | 0.6703 |
| SAFF (Ours) | | √ | 0.6745 |
| **SAFF (Ours)** | **√** | **√** | **0.6863** |

表 2 PASCAL VOC 2012 验证集前景分割 F-测度比较

| 方法 | 样本均衡 | F-测度 |
|---|---|---|
| DRFI [16] | | 0.6388 |
| CAM [14] | | 0.5828 |
| SAFF (Ours) | | 0.6592 |
| **SAFF (Ours)** | **√** | **0.6725** |

结果更好地指导了分割网络对前景提取的能力和在新数据上的泛化能力。

图 2 展示了不同方法在准确率-召回率曲线对比。除了基线以外，研究同样比较了在样本均衡、分割网络重推理策略使用与否条件下的性能结果。观察到，基于训练集、验证集的准

图 3 展示了一些可视化结果，可以看出，研究提出的方法得到的前景分割同时收获了更好的语义区域定位和完整前景区域挖掘的性能。特别的，第 4 行的结果展示了通过图间多实例学习，初始的错误前景分割种子仍可以被修正，并最终得到高质量的前景区域提取结果。

## 6　结论

在本研究中，一种基于表观-语义特征融合的无监督前景分割方法被提出。对于一张图像，研究中首先计算了图中每个超像素的一元区域语义和表观特征。接着，通过交叉地利用表观和语义特征的相似度量，实现了两种特征的融合，并计算得到了每个区域的二元图内上下文信息。进一步的，自适应权重学习的策略被建立，通过自动调节每一个特定图像实例中，各个维度特征对前景估计的影响，一种最优前景分割的加权参数被得到，并实现了对图像前景置信度的推理。最后，分割网络模型被使用来

学习不同实例之间的前景共性特征。利用训练得到的网络模型，图像可以重推理得到更准确的前景分割结果。在 PASCAL VOC 2012 训练集和测试集上的实验分别证明了算法的有效性和泛化能力。同时，本研究提出的方法可将其他前景分割方法作为基线，并广泛应用于对前景分割、弱监督语义分割等任务性能的提升。

研究认为，在未来工作中，考虑引入多种不同类型语义、表观特征融合，并采用交替迭代的方式，挖掘图像内部空间上下文信息和不同图像间的共性表达特征，是实现更精准前景分割和语义分割的重要思路。

# 7 致谢

# 8 参考文献